\documentclass{article}
\usepackage{spconf,amsmath,amssymb,graphicx,booktabs}
\usepackage{url}

\usepackage{enumitem}
\setlist{nosep, leftmargin=14pt}

\usepackage{mwe} 

\title{When Annotation Patterns Are Learned: Skin-Tone Dependence in Dermatology Concepts}
\name{Yijian Li, Mei~Bigliardi-Qi, Saad Bedros, Paul~Bigliardi, Nikolaos~Papanikolopoulos$^{\star}$}
\address{University of Minnesota \qquad $^{\star}$Corresponding author: papan001@umn.edu}
\begin{document}
%
\maketitle
\begin{abstract}
Concept bottleneck models (CBMs) let clinicians inspect and correct skin disease diagnoses through clinical concepts, but only if the concept annotations are reliable. We therefore audit the reliability of SkinCon, a widely used dermatology concept dataset, across skin tones. Within the same disease and with near duplicates removed, erythema is annotated far less often on dark than on light skin, with an odds ratio of 0.17. In contrast, hyperpigmentation and hypopigmentation are annotated more often, with odds ratios of 6.2 and 4.5, respectively. We refer to this pattern as tone-dependent concept annotation. An evidential concept model trained on these annotations learns the same pattern and reproduces it on unseen images. Such a model may agree with potential erythema omissions rather than contradict them. Consistent with this, three label error detectors underperform a disease-only baseline at finding erythema that dermatologist captions mention but SkinCon labels as absent. The three detectors are based on confident learning, training dynamics, and the model's own evidential uncertainty. Concept annotations should thus be audited within each subgroup before explanations built on them can be trusted.
\end{abstract}


\section{Introduction}
\label{ch:intro}
Skin tone can systematically bias clinical assessment~\cite{BenGashir02}. For example, on darkly pigmented skin, the contrast between red and green that makes erythema visible is reduced. The surrounding pigment also often masks the redness. In 2021, the British Association of Dermatologists acknowledged this problem by issuing guidance on describing erythema in skin of color~\cite{BAD21}. Because dermatology concept datasets are annotated from photographs, a sign that is difficult for a clinician to see at the bedside is likely also difficult for an annotator to identify in an image.

Concept bottleneck models (CBMs)~\cite{Koh20,Espinosa22} are increasingly used for interpretable skin disease diagnosis~\cite{Gao24}. They route each diagnosis through an intermediate layer of clinical concepts defined by experts. This makes the model's reasoning readable and correctable. A clinician who disagrees with a predicted concept, such as ``plaque present'', can intervene on it. Both properties assume that the concept annotations are correct. If a concept is systematically misannotated for a subgroup, both fail together. The explanation remains faithful to the model, but the model's reasoning rests on incorrect labels. Intervention, in turn, relies on a clinician who may be equally unable to see the error in the photograph.

Label error detectors~\cite{Northcutt21,Pleiss20} and evidential concept models~\cite{Gao24} are designed to flag unreliable labels and concepts. However, both rely on a model trained on the very annotations they are meant to check. Confident learning, for instance, flags a label when the model disagrees with it. It assumes that label noise depends only on the true class. An omission caused by poor visibility violates this assumption because the sign is missed precisely in images where it is faint. A model trained on these labels may then learn to reproduce the omission rather than disagree with it, and the error may go undetected. To our knowledge, label noise that depends on group membership has been modeled for fair classification~\cite{Wang21} but not for concept annotation.

We make three contributions. First, within the same disease, SkinCon~\cite{Daneshjou22} annotates erythema less often on dark skin and pigmentary changes more often. We call this tone-dependent concept annotation. Second, a concept model trained on these annotations reproduces it on unseen images. Third, consistent with this, three detectors underperform a disease-only baseline at finding potential erythema omissions.

\section{Methods}
\subsection{Data}
Fitzpatrick17k~\cite{Groh21} is a public dataset of 16,577 clinical dermatology images covering 114 skin conditions. Of these, 3,230 carry SkinCon~\cite{Daneshjou22} annotations. These describe each image with 48 binary clinical concepts defined by dermatologists. Following~\cite{Gao24}, we retain the 22 concepts with more than 50 positive examples. Nearly all images also have a Fitzpatrick skin type (FST) label. FST I and II are grouped as light, III and IV as mid, and V and VI as dark. The annotated set includes 1,524 light, 1,198 mid, and 424 dark images. The other 84 lack an FST label.

Fitzpatrick17k contains many near-duplicate images and inconsistent FST labels, as shown by an independent audit~\cite{Abhishek25}. We therefore remove duplicates using the audit's keep list, retaining 11,394 of the 16,577 images. Of the 3,230 annotated images, 2,217 remain. They include 1,176 light, 743 mid, and 242 dark images. The other 56 lack an FST label. The odds ratio analysis and concept model training are run on both sets, with the deduplicated set as primary. Because the keep list does not correct FST labels, we also check the odds ratios using the second set of FST labels in Fitzpatrick17k, from Centaur Labs~\cite{Groh21}. The first set, from Scale AI, remains primary.

\subsection{Quantifying annotation dependence on skin tone}
For each concept $C_k$, we model the annotation of image $i$ by a logistic regression with disease fixed effects:
\begin{equation}
\operatorname{logit} P(C_{ik}=1) = \alpha_{k,d(i)} + \beta_k D_i ,
\label{eq:or}
\end{equation}
where $C_{ik}\in\{0,1\}$ is the SkinCon annotation of $C_k$ on image $i$, $d(i)$ is the disease of image $i$ with intercept $\alpha_{k,d(i)}$, $D_i$ equals 1 for dark skin and 0 for light skin, and $\beta_k$ is the shift in log odds for dark skin, shared across diseases. Only light and dark images are used. The odds ratio of $C_k$ is then
\begin{equation}
\mathrm{OR}_k = e^{\beta_k} = \frac{\mathrm{odds}(C_{ik}=1 \mid D_i=1,\ d(i)=d)}{\mathrm{odds}(C_{ik}=1 \mid D_i=0,\ d(i)=d)},
\label{eq:ratio}
\end{equation}
where $\mathrm{odds}(A)=P(A)/(1-P(A))$. Because the ratio is defined within each disease $d$, differences in disease composition between tone groups do not affect $\mathrm{OR}_k$. The fit includes only diseases in $\mathcal{D}_k=\{d:\ n_d\ge 10,\ 0<\bar D_d<1,\ 0<\bar C_{dk}<1\}$, where $n_d$ is the number of light and dark skin images of disease $d$, and $\bar D_d$ and $\bar C_{dk}$ are the means of $D_i$ and $C_{ik}$ within $d$. The last two conditions require both tone groups and both annotation values to be present; diseases that violate them carry no information about $\beta_k$. The 95\% percentile interval of $\mathrm{OR}_k$ comes from 2,000 bootstrap resamples of images. Each resample repeats the disease selection and the fit.

To test whether the concept model inherits these dependencies, we refit Eq.~\eqref{eq:or} with the predicted probability in place of $C_{ik}$:
\begin{equation}
\operatorname{logit} \mathbb{E}[\bar p_{ik} \mid d(i), D_i] = \alpha'_{k,d(i)} + \beta'_k D_i ,
\label{eq:model}
\end{equation}
where $\bar p_{ik}$ is $p_{ik}$ from Section~\ref{ch:signals}, averaged over the three seeds. This fractional logistic regression uses the same disease sets $\mathcal{D}_k$ and bootstrap, and its odds ratio is $e^{\beta'_k}$. As controls, both fits are repeated for papule and scale, two morphological concepts that do not depend on color.

\subsection{Detection protocol and baseline}
\label{sec:protocol}
For concept $k$, we evaluate images that SkinCon labels as absent and that have an English SkinCAP caption~\cite{Zhou24}. GPT-4 refined and translated the captions, and dermatologists checked them. Each caption sentence is searched for terms that indicate concept $k$. Negated mentions such as ``no erythema'' are ignored. For erythema, the terms include ``erythematous'' and ``red''. ``Violaceous'' is added because erythema often looks violaceous on dark skin~\cite{BAD21}. If a term appears in a sentence about the image itself rather than the disease in general, the image is a positive and a potential omission. Images without a mention count as negatives, but the concept may still be present.

Because a signal could perform well by exploiting the disease mix alone, we compare every signal with a disease-only baseline that uses no image information. The baseline scores image $i$ by $r_{k,d(i)}$, the fraction of annotated images of disease $d(i)$ that SkinCon labels with concept $k$. It uses SkinCon annotations only, so it has no access to the SkinCAP reference. It is evaluated on the same images as the signals. Performance is measured by AUC, with paired 95\% intervals for each signal's difference from the baseline over 2,000 bootstrap resamples.

\begin{figure*}[t]
  \centering
  \includegraphics[width=\textwidth]{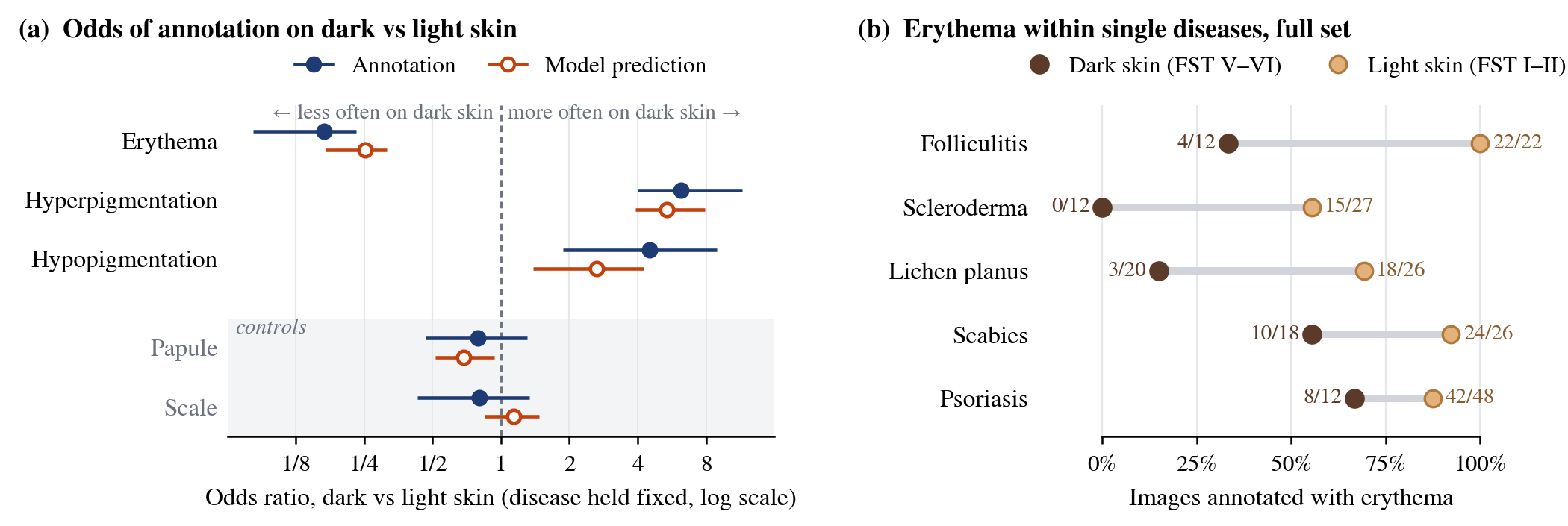}
  \caption{(a) Odds ratio of each concept on dark vs light skin for the annotations (filled) and the model's predictions (open) in the deduplicated set, with 95\% bootstrap intervals; papule and scale are controls. (b) Erythema annotation rate on dark and light skin in the full annotated set, shown for illustration only. The five diseases have the largest drop from light to dark among those with at least 10 images in each tone group.}
  \label{fig:bias}
\end{figure*}

\subsection{Signals under test}
\label{ch:signals}
We evaluate one signal from each of three detector families. For confident learning~\cite{Northcutt21}, we use the self-confidence score $q_{ik}=p_{ik}^{C_{ik}}(1-p_{ik})^{1-C_{ik}}$, where $p_{ik}$ is the predicted probability of concept $k$ for image $i$ from a model trained without image $i$. For training dynamics, the signal is the area under the margin (AUM)~\cite{Pleiss20}, computed on predicted probabilities rather than logits. It averages the margin $(2C_{ik}-1)(2p^{(t)}_{ik}-1)$ over the training epochs $t$ in which image $i$ is included, where $p^{(t)}_{ik}$ is the prediction at epoch $t$. For the model's own uncertainty, the signal is evidential uncertainty $u_{ik}=2/(a_{ik}+b_{ik})$~\cite{Gao24}, where $a_{ik}$ and $b_{ik}$ are the parameters of the predicted Beta distribution with mean $p_{ik}$.  A low $q$, a low AUM, or a high $u$ indicates a suspected error.

All three signals require a trained concept model. We use evi-CEM~\cite{Gao24}, an evidential concept embedding model, without its concept rectification step. It uses a ResNet-34 backbone and predicts one of three coarse categories: benign, malignant, or nonneoplastic. We train it separately on the 2,217 deduplicated and 3,230 full annotated images, using five-fold cross-validation and three random seeds.

\section{Results}
\subsection{Annotation depends on skin tone in both directions}
Within the same disease, SkinCon annotates erythema far less often on dark skin and both pigmentary concepts more often. In Fig.~\ref{fig:bias}(a), the odds ratio is 0.17 [0.08, 0.23] for erythema, 6.19 for hyperpigmentation, and 4.50 for hypopigmentation. The two control concepts show no evidence of a tone difference. The pattern holds on the full set, with odds ratios of 0.16, 5.57, and 3.55. With the Centaur Labs labels, they are 0.14, 9.93, and 6.28 on the deduplicated set and 0.15, 8.03, and 4.23 on the full set. None of these 12 intervals includes 1. Fig.~\ref{fig:bias}(b) shows the same erythema pattern within individual diseases. If the erythema difference reflects omissions, images that SkinCon labels as absent should more often have erythema on dark than on light skin. Their SkinCAP captions should then mention it more often on dark skin, giving a disease-adjusted odds ratio above 1. This could not be confirmed. On the full set, this SkinCAP odds ratio is 0.93 [0.16, 1.81], from only 109 images across seven diseases.

\begin{table}[t]
\centering
\caption{Detection AUC on the deduplicated set. pos: potential omissions. Totals include mid and unlabeled tones. Detector values are means over three seeds. Baseline: disease-only; $u$: evidential uncertainty; CL: self-confidence $q$.}
\label{tab:detect}
\small
\setlength{\tabcolsep}{4pt}
\begin{tabular}{lccccc}
\toprule
 & pos / $n$ & Baseline & $u$ & AUM & CL \\
\midrule
Erythema         & 188 / 643 & 0.723 & 0.483 & 0.646 & 0.633 \\
\quad dark skin  & 37 / 121  & 0.618 & 0.455 & 0.517 & 0.474 \\
\quad light skin & 83 / 273  & 0.760 & 0.486 & 0.660 & 0.660 \\
\midrule
Hyperpig.        & 42 / 1604 & 0.749 & 0.664 & 0.751 & 0.672 \\
\bottomrule
\end{tabular}
\end{table}

\begin{table}[t]
\centering
\caption{Confident learning flag rate on dark relative to light skin,
deduplicated set. Ranges and ranks are over three seeds, among 22 concepts.
Rows are the top three by rank and the two concepts specified in advance
($^\dagger$). Sig.: seeds with a ratio above 1 and $p<0.05/22$ in a
two-proportion $z$-test.}
\label{tab:cleanlab}
\small
\setlength{\tabcolsep}{5pt}
\begin{tabular}{lccc}
\toprule
Concept & Dark\,/\,light & Rank & Sig. \\
\midrule
Hypopigmentation            & 3.09--3.38 & 1\,/\,1\,/\,1  & 3/3 \\
Scar                        & 2.36--2.68 & 2\,/\,2\,/\,2  & 1/3 \\
Patch                       & 2.10--2.32 & 3\,/\,3\,/\,3  & 2/3 \\
Hyperpigmentation$^\dagger$ & 1.66--1.77 & 4\,/\,4\,/\,5  & 3/3 \\
Erythema$^\dagger$          & 1.33--1.53 & 11\,/\,6\,/\,9 & 0/3 \\
\bottomrule
\end{tabular}
\end{table}

\subsection{The model inherits the pattern}
\label{sec:inherit}
Fig.~\ref{fig:bias}(a) shows that the model's predictions on unseen images reproduce the pattern for all three concepts. For the two controls, the model's odds ratios do not differ clearly from those of the annotations. The model is never given skin type, and each odds ratio compares images of the same disease. It must therefore draw the pattern from image content, possibly skin color itself. For erythema, the effect is smaller than in the annotations. The model's odds ratio is 0.25 [0.17, 0.32], versus 0.17 for the annotations. Each seed alone gives 0.22 to 0.28, and the full set gives 0.26 versus 0.16. The smaller effect may partly reflect scale, because the model outputs probabilities while the annotations are binary. As a sensitivity check, we binarize each concept at the threshold where it is predicted as often as it is annotated. The three odds ratios become 0.17, 8.94, and 4.37. These effects are about as strong as in the annotations, or stronger.

\subsection{Three detectors trail the baseline on erythema}
If the model has learned this pattern, its signals may fail to rank potential omissions above other images. Table~\ref{tab:detect} measures this by AUC for two concepts specified in advance. For erythema, all three signals score below the disease-only baseline. With an AUC of 0.48, $u$ is no better than chance. Paired differences average $-0.24$ for $u$, $-0.08$ for AUM, and $-0.09$ for CL. Every 95\% interval lies below zero in all three seeds. The highest upper bounds are $-0.17$ for $u$, $-0.02$ for AUM, and $-0.03$ for CL. The result holds even when the baseline is computed with five-fold cross-validation. This lowers its AUC only slightly to 0.718. Each skin tone group shows the same direction, but the groups are small and some intervals cross zero. On the full set, all three erythema estimates also fall below the baseline. For hyperpigmentation, no signal clearly outperforms the baseline. On the full set, AUM exceeds it mainly because the baseline falls from 0.749 to 0.721.

\subsection{Detector flags follow the annotation pattern}
\label{sec:flags}
In an exploratory analysis, we ask whether an auditor could detect this pattern from detector flags alone. Confident learning with cleanlab's default settings checks each SkinCon annotation and flags those that are potentially incorrect. SkinCAP is not used here. Table~\ref{tab:cleanlab} ranks the 22 concepts by the ratio of their flag rates on dark versus light skin. The two pigmentary concepts rank near the top and remain significant after Bonferroni correction. Erythema is never significant and would be overlooked. Each flag also has a direction, because a flag on a ``present'' annotation suggests removing it and a flag on an ``absent'' annotation suggests adding it. Separate rates for present and absent annotations show that the modest overall ratio of erythema mixes two opposite trends. Erythema marked present is flagged 2.98--4.33 times as often on dark as on light skin, while erythema marked absent is flagged only 0.41--0.55 times as often. Hyperpigmentation marked absent is flagged 2.53--3.75 times as often. In each case, the flags point in the same direction as the annotation pattern, with or without adjustment for disease.

\section{Discussion}
CBMs assume that concept annotations are reliable. Our results show that tone-dependent patterns in SkinCon annotations pass into a model trained on them. Potential erythema omissions also slip past the three detectors tested here. Both findings are consistent with one explanation. A miss that the model has learned may look correct to detectors built on that model.

The erythema pattern and the pigmentary patterns point in opposite directions and may have different causes. Erythema appears less red on dark skin~\cite{BAD21}, so its pattern most plausibly reflects what annotators can see. In contrast, pigmentary changes stand out more on dark skin and are more expected there. They are also more common after inflammation in darker skin~\cite{Davis10}, so their pattern may be partly real. The SkinCAP check cannot settle this because its dermatologists saw the same photographs and may have missed the same signs. Whatever the cause, the model learns the pattern, and the detectors cannot tell whether a difference is real or comes from annotation.

One possible mechanism fits the model and detector findings. Confident learning assumes that label noise depends only on the true class. Erythema missed because it looks faint would break this assumption. Such image-dependent noise is known to be hard to handle~\cite{Xia20,Chen21}. The model may learn such omissions from the image and agree with those labels. The disagreement-based signals CL and AUM would miss many of them. Evidential uncertainty only measures how much evidence the model has and ignores whether it disagrees with the label. This is consistent with all three falling behind the disease-only baseline. Faint erythema can also be missed on light skin, where the detectors fall short too.

CBMs are often trained jointly on concepts and diagnosis, so concept labels shape both what the network learns and how it explains itself. A CBM trained on SkinCon may therefore show erythema less often in its explanations on dark skin, as the annotations do. Our concept model already does this, with an odds ratio of 0.25 for erythema. A clinician who intervenes on these concepts from the photograph alone may miss the sign too. Concept accuracy measured against these labels could also overstate performance on dark skin. Other concept datasets annotated from photographs may show the same tone dependence. To detect and reduce it, we suggest three steps. First, report disease-adjusted odds ratios by skin tone for every concept. Second, compare any label error detector with a disease-only baseline. Third, describe color signs in annotation guidelines as they look on dark skin, where erythema often appears violaceous or dusky rather than red~\cite{BAD21}.

\noindent\textbf{Limitations.} First, our audit identifies which concepts are annotated differently on dark and light skin, but not which images are mislabeled. Second, the evidence comes from one dataset with few dark-skin images, so some intervals are wide. Images of the same disease may still differ in stage, treatment, body site, or photo conditions. FST is only a noisy proxy for skin tone~\cite{Abhishek25}, but the odds ratios hold under a second set of labels. Third, all model and detector results come from one concept model. Fourth, the reference itself may favor the disease-only baseline. The SkinCAP dermatologists proposed diagnoses themselves, so they may have mentioned erythema partly because it is expected for that disease rather than clearly visible in the image. We exclude general statements about the disease, but the diagnosis may still have influenced the descriptions. A natural next step is to extend the audit to Diverse Dermatology Images (DDI)~\cite{Daneshjou22b}, because SkinCon annotates its images with the same 48 concepts.

\section{Conclusion}
We find tone-dependent concept annotation in SkinCon. Within the same disease, erythema labels are less frequent on dark skin and pigmentary labels more frequent, with an erythema odds ratio of 0.17. A concept model inherits this pattern, with an erythema odds ratio of 0.25. Consistent with this, three detectors built on the model trail a disease-only baseline in finding potential erythema omissions. Concept annotations should be audited within each subgroup, with independent review of suspected errors.

\section{Compliance with Ethical Standards}
This study is a retrospective analysis of open-access data from Fitzpatrick17k~\cite{Groh21}, SkinCon~\cite{Daneshjou22}, and SkinCAP~\cite{Zhou24}. No new data were collected. Ethical approval was not required for this use of public data.

\section{Acknowledgments}
No funding was received for this study. The authors have no relevant financial or non-financial interests to disclose. The authors thank the Minnesota Supercomputing Institute at the University of Minnesota for computing resources. Anthropic's Claude was used only to edit the language of text written by the authors. It did not contribute scientific content. All edits were reviewed by the authors.

\bibliographystyle{IEEEbib}
\bibliography{refs}

\end{document}